# Hybrid GCN–GRU Model for Anomaly Detection in Cryptocurrency Transactions

Gyuyeon Na[1], Minjung Park[2], Hyeonjeong Cha[1], Soyoun Kim[1], Sunyoung Moon[1], Sua Lee[1], Jaeyoung Choi[1], Hyemin Lee[1], and Sangmi Chai[1,3*]

[1] AI and Business Analytics, Ewha Womans University, Seoul, Republic of Korea
amy-na@ewha.ac.kr, hyeonjeong.cha@ewha.ac.kr, sykim07@ewha.ac.kr,
sunyoung.moon427@ewha.ac.kr, sua724@ewha.ac.kr,
lovejesus021114@gmail.com, gpals5371@ewha.ac.kr, hyemin.lee@ewha.ac.kr
[2] Department of Business Administration, Kumoh National Institute of Technology, Gumi, Republic of Korea
mjpark@kumoh.ac.kr
[3] Coretrustlink, Seoul, Republic of Korea

**Abstract.** Blockchain transaction networks are complex, with evolving temporal patterns and inter-node relationships. To detect illicit activities, we propose a hybrid GCN–GRU model that captures both structural and sequential features. Using real Bitcoin transaction data (2020–2024), our model achieved 0.9470 Accuracy and 0.9807 AUC-ROC, outperforming all baselines.

## 1 Introduction

Anomaly detection in blockchain transaction networks has emerged as a critical research area. This is driven by the rapid growth of cryptocurrency markets and the increasing prevalence of illicit activities, such as money laundering and transaction mixing. Detecting such anomalous transactions is particularly challenging because they are often concealed within large volumes of legitimate transactions and exhibit complex patterns that span both the network topology and temporal dynamics. Traditional machine learning models, including ensemble-based approaches such as Random Forest, have demonstrated strong performance in various classification tasks by aggregating multiple weak learners. However, these models typically rely on feature vectors that do not explicitly capture the underlying structural relationships between entities in a transaction network or the sequential patterns that evolve over time. As a result, their ability to generalize to previously unseen patterns, especially in dynamically evolving blockchain networks, is limited. Graph-based models, such as Graph Convolutional Networks (GCNs), have shown promise in learning structural dependencies by leveraging the connectivity of transaction graphs. Similarly, sequence models, such

---

* Corresponding author: smchai@ewha.ac.kr

Gated Recurrent Units (GRUs), excel at modeling temporal dependencies in sequential data. Nonetheless, when applied individually, these models only capture one aspect of the problem—either the spatial (structural) domain or the temporal domain—thus potentially missing critical cross-domain patterns that may be indicative of anomalous behavior. The primary objective of this study is to investigate the performance improvement achieved by jointly modeling network-based relational structures and time-series behavioral patterns in the context of anomalous transaction detection. To this end, we propose and evaluate a hybrid deep learning architecture that integrates GCN and GRU components, enabling the model to learn both spatial and temporal patterns simultaneously. This dual-perspective approach is expected to enhance detection accuracy, particularly in scenarios where anomalous behavior manifests through subtle interactions between structural and temporal cues. The contributions of this work are as follows.

1. We provide a systematic comparison of baseline models, including Random Forest, CNN, and GRU, under identical preprocessing and feature extraction pipelines.
2. We propose a hybrid GCN–GRU architecture designed to capture both structural and temporal dependencies in blockchain transaction data.
3. We conduct comprehensive experiments on real-world Bitcoin mixing transaction datasets, evaluating performance across multiple metrics and analyzing the impact of combining structural and temporal features.

## 2   Related Works

### 2.1   Anomaly Detection in Cryptocurrency Markets

With the recent surge in Bitcoin prices, trading volume has increased significantly, leading to a rise in illicit transactions. The amount of funds funneled into illicit cryptocurrency addresses reached 37.4 billion USD in 2021, 56.6 billion USD in 2022, 58.7 billion USD in 2023, and 44.7 billion USD in 2024 [1]. Furthermore, a report published in December 2024 indicated that cryptocurrency platforms suffered approximately USD 2.2 billion in hacking losses, further illustrating the growing urgency of monitoring and detecting suspicious activities in the cryptocurrency market [2]. As criminal methods evolve, mixing services have become a widely used tactic to sever traceable links between transaction senders and recipients. Identified 19 different types of mixing services, including centralized, decentralized, cross-chain, and cryptocurrency-based services [3][4]. The need for anomaly detection techniques tailored to blockchain and cryptocurrency environments is emerging. The fundamental characteristics of blockchain are immutability and decentralization. Once a transaction is recorded, it cannot be altered or deleted, and since no central authority exists, no entity has the power to immediately block or reverse a transaction. Although these structural constraints are useful for retrospective analysis, they impose inherent limitations on the ability to detect and respond to anomalous activities in real time [5]. Mixing services actively exploit these general features of the blockchain. By merging multiple transactions and obfuscating address chains, they transform the flow of funds from a one-to-one mapping into many-to-many mappings [7]. As a result, causal connectivity within the transaction graph is weakened and traceability is significantly reduced. Consequently, the combination of blockchain immutability and decentralization – which prevent intervention –with the obfuscation and structural complexity introduced by mixing services makes real-time anomaly detection extremely challenging [6]. To overcome the aforementioned limitations, some studies have sought to address the problem by leveraging machine learning and data mining techniques in greater depth. In the Bitcoin transaction network, unsupervised learning methods were employed to detect users and transactions suspected of abnormal activities. K-means, Mahalanobis distance, and unsupervised SVM (v-SVM) algorithms were applied; however, the performance metrics obtained were relatively poor [5]. In addition, for the detection of anomalies in Bitcoin transaction data, the unsupervised DBSCAN algorithm was applied to perform density-based clustering, identifying various types of clusters and classifying noise points as anomalies. However, the evaluation results, precision: 1.0, recall: 0.0068, F1 score: 0.0135, and AUC-ROC: 0.5034, indicate that the method was ineffective in accurately detecting anomalies [8]. Together, these works highlight the pressing need for research on real-time anomaly detection in cryptocurrency markets, emphasizing approaches that not only ensure technical robustness, but also maintain adaptability, interpretability, and resilience against the rapidly evolving landscape of illicit activities. Despite notable advances, most existing studies have relied on supervised learning methods applied to retrospective analyzes of past transaction data. As a result, these approaches are inherently constrained as they cannot provide timely detection of anomalies while transactions are unfolding in real time. Consequently, future research should prioritize the development of scalable techniques capable of operating in dynamic, unsupervised environments, enabling the immediate identification and mitigation of illicit activities before they propagate further within the financial ecosystem.

### 2.2 Deep Learning-based Approaches and Limitations

Bitcoin financial transactions can be effectively represented as graph structures that capture the flow of value between addresses. These transaction graphs have been widely used in prior research, where each transaction is modeled as a node and enriched with associated features for training machine learning and deep learning models. This graph-based approach enables the analysis of both individual transaction attributes and the structural interactions among them, thereby facilitating deeper pattern discovery and improving the accuracy of fraud detection [9]. In addition, the integration of sequential learning models has further advanced fraud detection. For example, employing a Gated Recurrent Unit (GRU) to capture temporal dependencies, followed by a Random Forest classifier, has been shown to enhance both stability and accuracy in detecting fraudulent activity [10]. Such methods leverage the strengths of recurrent neural networks in modeling sequential patterns while benefiting from ensemble learning for robust prediction. More recently, graph-based deep learning models have been extensively applied across diverse domains—including social network analysis, fraud detection, traffic forecasting, and computer vision—owing to their flexibility and strong representational power. Convolutional Neural Networks (CNNs), well known for their ability to process complex data, have been successfully combined with graph structures in Graph Convolutional Networks (GCNs), yielding promising results in tasks such as classification, labeling, and link prediction. Applied to transaction networks, GCNs have demonstrated significant improvements in fraud detection performance [9]. However, while GCNs excel at capturing structural properties of networks, they struggle to fully account for temporal continuity and sequential anomaly patterns in financial transaction data. Conversely, GRUs are effective in learning temporal dependencies but cannot directly encode graph structures. To address these complementary limitations, recent research has introduced hybrid models that combine GCNs with GRUs. By leveraging GCNs to learn structural relationships among nodes and GRUs to capture temporal continuity and dynamic variations, these models can effectively represent long-term dependencies in transaction networks, ultimately achieving superior fraud detection performance [12]. In summary, while previous studies have incorporated both structural properties of networks and temporal continuity, the highly dynamic and rapidly evolving patterns of Bitcoin transactions necessitate real-time anomaly detection techniques. Such an approach enables the immediate identification of illicit activities, thereby allowing for faster and more effective countermeasures.

## 3 Materials and Methods

In this section, we present a comprehensive experimental framework for detecting anomalous cryptocurrency transactions using hybrid graph–sequence model. The procedure consists of four stages: 1. data collection and preprocessing, 2. baseline model implementation, 3. proposed model architecture, and 4. experimental setup and evaluation, as detailed in subsections 3.1 3.4.

### 3.1 Data Collection and Preprocessing

The dataset used in this study comprises Bitcoin mixing transactions via the Wasabi Wallet between January 1, 2020, and April 24, 2024. Each file corresponds to a cluster of cryptocurrency addresses owned by the same entity. The *Counterparty Address* denotes the representative address of a partner's cluster, which may differ from the actual transaction address.

During data cleaning, missing values were removed (*Date*, *Receiving Address*, or *Counterparty Address*). While extreme values (e.g., large amounts) were retained to preserve potential anomalies. After cleaning, the dataset contained 318,388 normal (label = 0) and 69,031 anomalous transactions (label = 1).

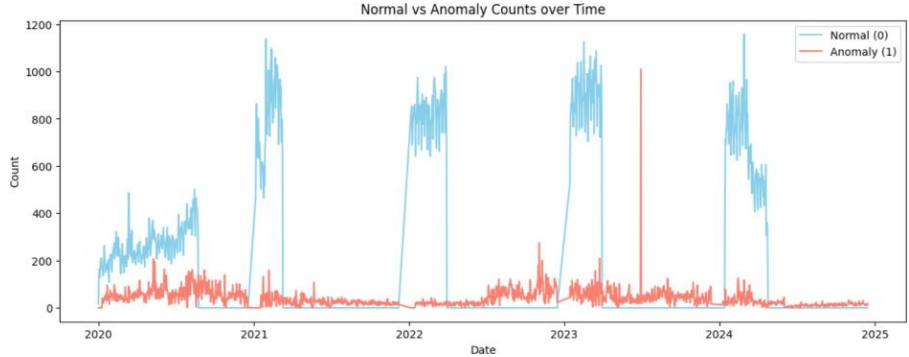

**Fig. 1.** Normal vs Anomaly Counts over Time

**Table 1.** Structure of the anomalous transaction dataset.

| Column | Description |
|---|---|
| Hash | Unique transaction identifier used to reconstruct the full transaction topology from the blockchain. |
| Date (UTC) | Transaction timestamp in Coordinated Universal Time. |
| Receiving Address | Address within the subject cluster that receives cryptocurrency. In Bitcoin, Value can be negative due to net amount calculations involving hidden inputs/outputs. |
| Counterparty Address | Representative address of the counterparty cluster; may differ from the actual transaction address involved. |
| Counterparty Cluster Name | Official name of the counterparty cluster. |
| Counterparty Shared Name | Private or informal name of the counterparty cluster (optional). |
| Counterparty Category | Category of the counterparty cluster (e.g., exchange, mixing service). |
| Value | Cryptocurrency amount; positive for inbound transactions, negative for outbound transactions (net). |
| USD Value | USD-equivalent amount at the time of the transaction. |

For feature engineering, the *Date* field was converted to UTC, with *Hour* and *DayOfWeek* derived as temporal features. Numerical attributes such as *Value* and *USD Value* were normalized to [0,1] using MinMaxScaler. The dataset was split chronologically. Transactions from 2020–2022 for training and 2023–2024 for testing, ensuring no temporal leakage. Longer-term seasonal features (e.g., *Month*, *Quarter*) were excluded to focus on short-term temporal dynamics.

### 3.2 Baseline Model Implementation

Several baseline models were implemented for comparative evaluation. The Random Forest model , a classical ensemble learning method based on bootstrap aggregation was chosen for its robustness to noisy data and capability to capture complex non-linear relationships in tabular features. The Graph Convolutional Network (GCN), a spatial graph neural network, was employed to propagate and aggregate information across a feature correlation graph constructed from statistical dependencies among features. The Convolutional Neural Network (CNN), specifically a one-dimensional CNN, was applied over sliding windows to extract local temporal patterns, enabling kernel filters to learn representations from short-term sequential segments. The GCN–CNN hybrid architecture was designed where GCN layers first enhanced each time step's features followed by CNN layers to capture temporal dependencies more effectively.

Finally, the Gated Recurrent Unit (GRU) was selected. GRU is a recurrent neural network designed to capture long-term temporal dependencies and to mitigate vanishing-gradient issues. Compared with LSTM, it has a lower computational cost. It still delivers competitive performance.

Table 2 summarizes the key hyperparameters and preprocessing procedures for each baseline model. The Random Forest model used a temporal feature set derived from datetime expansion, with median imputation applied only to handle missing values introduced during this feature engineering process. The GCN model required graph-specific preprocessing, including feature correlation graph construction and node feature standardization. The CNN model shared the same temporal feature set and preprocessing as Random Forest. The GCN–CNN hybrid combined both spatial and temporal processing by applying GCN layers at each time step followed by CNN layers. Finally, the GRU model also used the same temporal feature set and preprocessing as Random Forest.

**Table 2.** Baseline model hyperparameters and preprocessing summary

| Model | Hyperparameters | Preprocessing / Features |
|---|---|---|
| RandomForest | n_estimators=200, random_state=42 | Datetime expansion (year, month, day, dow, hour), median imputation |
| GCN | hidden_dim=64, epochs=50, lr=0.01, weight_decay= $5 \times 10^{-4}$ | Graph from feature correlations, node feature standardization |
| CNN | filters=64, kernel=3, activation=ReLU | Sliding window, same temporal features as RF, reshaped for CNN |
| GCN–CNN | GCN hidden_dim=64, CNN filters=64, kernel=3 | GCN at each timestep, then CNN for temporal deps, graph+seq input |
| GRU | hidden_size=64, epochs=10, batch=64, Adam | Sliding window, same temporal features as RF, reshaped to (samples, timesteps, features) |

### 3.3 Proposed Model Architecture

Fig. 2 illustrates the overall GCN–GRU minimal pipeline, where per-step graph convolution captures spatial dependencies and the GRU encodes temporal dynamics before classification.

The proposed architecture combines GCN and GRU to jointly model the structural dependencies and temporal dynamics of blockchain transaction networks.

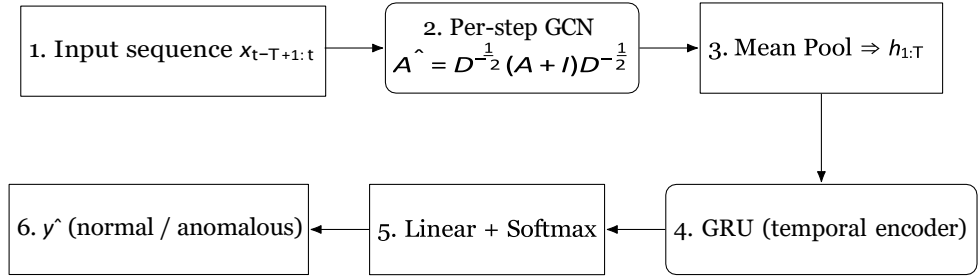

**Fig. 2.** GCN–GRU (minimal pipeline)

*Graph Convolutional Layer.* Let $X \in R^{N \times F}$ be the node feature matrix, where $N$ denotes the number of transactions and $F$ the feature dimension, and let $A \in R^{N \times N}$ be the adjacency matrix. The GCN layer updates the hidden representation $H^{(l)}$ according to:

$$H^{(l+1)} = \sigma\left(\hat{A} H^{(l)} W^{(l)}\right),$$

where $\hat{A} = D^{-\frac{1}{2}}(A + I)D^{-\frac{1}{2}}$ is the symmetrically normalized adjacency matrix with self-loops, $W^{(l)}$ is the learnable weight matrix, and $\sigma(\cdot)$ is a non-linear activation function (e.g., ReLU).

*Gated Recurrent Unit Layer.* The output embeddings from the GCN are temporally ordered according to transaction timestamps, producing a sequence $\{x_t\}_{t=1}^{T}$. The GRU cell computes:

$$z_t = \sigma(W_z x_t + U_z h_{t-1}), \quad r_t = \sigma(W_r x_t + U_r h_{t-1}),$$

$$\tilde{h}_t = \tanh(W_h x_t + U_h(r_t \odot h_{t-1})), \quad h_t = (1 - z_t) \odot h_{t-1} + z_t \odot \tilde{h}_t,$$

where $z_t$ and $r_t$ are the update and reset gates, respectively, and $\odot$ denotes element-wise multiplication.

*Hyperparameters.* Unless otherwise stated, the following hyperparameters are used: GCN hidden dimension $d_g = 64$, GRU hidden dimension $d_h = 64$, $k = 5$ nearest neighbors in the feature graph, correlation threshold $\tau = 0.2$, dropout rate 0.1, window size $T = 10$, and stride $s = 1$.

*Output Layer.* The final hidden state $h_T$ is passed to a fully connected layer with softmax activation to yield the probability distribution over the target classes(illicit or legitimate).

*Design Rationale.* The GCN extracts spatial features from the transaction graph, capturing local structural dependencies, while the GRU models the sequential progression of transactions over time. This hybrid design enables the model to learn both topological and temporal patterns, improving its ability to detect anomalous blockchain activities

### 3.4 Experimental Setup and Evaluation

1) Experimental Setup

All models were implemented in PyTorch and trained in a CUDA-enabled GPU environment using the Adam optimizer with a learning rate of $1 \times 10^{-3}$, a batch size of 256, and a maximum of 30 epochs. Sliding windows of length 10 with a stride of 1 were applied, and the label of each sequence was determined by the final time step.

2) Performance Metrics

We evaluated model performance using Accuracy, Precision, Recall, F1-score, and AUC-ROC. The AUC-ROC score was computed as the integral of the true positive rate (TPR) over the false positive rate (FPR) across all classification thresholds, providing a robust measure of the model's discriminative ability under both balanced and imbalanced data conditions. **Accuracy**: Measures the proportion of correctly classified instances among all instances.

$$\text{Accuracy} = \frac{TP + TN}{TP + TN + FP + FN}$$

**Precision**: Measures the proportion of correctly predicted positive instances among all predicted positives.

$$\text{Precision} = \frac{TP}{TP + FP}$$

**Recall**: Measures the proportion of correctly predicted positive instances among all actual positives.

$$\text{Recall} = \frac{TP}{TP + FN}$$

**F1-score**: Harmonic mean of Precision and Recall, providing a balance between the two.

$$\text{F1-score} = \frac{2 \times \text{Precision} \times \text{Recall}}{\text{Precision} + \text{Recall}}$$

**AUC-ROC**: Represents the area under the Receiver Operating Characteristic curve, reflecting the model's ability to discriminate between classes across all classification thresholds.

$$\text{AUC-ROC} = \int_0^1 TPR(FPR)\, d(\text{FPR})$$

where $TPR = \frac{TP}{TP+FN}$ and $FPR = \frac{FP}{FP+TN}$.

3) Reproducibility

The dataset was provided by Kloint. There is no expiration or restriction on its use. However, secondary or tertiary commercial use, such as resale by the research team, is prohibited. The dataset includes transactions from the Binance exchange, and transaction addresses were labeled as either normal or illicit based on transaction hashes identified as mixing-related. All implementation and model training were conducted in Python.

## 4 Results

Table 3 summarizes the evaluation results across Accuracy, Precision, Recall, F1-score, and AUC-ROC, highlighting clear differences among ensemble, structural, temporal, and hybrid models. Random Forest delivered strong Accuracy (0.9343) and AUC-ROC (0.9607), showing robustness to feature noise and complex decision boundaries. However, its relatively low Recall (0.6872) indicates a serious limitation in anomaly detection tasks where missed detections are costly. The GCN, focusing solely on graph topology, achieved moderate results (Accuracy = 0.7735, AUC-ROC = 0.6743), underscoring that static structural information alone cannot capture dynamic illicit behaviors. Temporal sequence models such as CNN and GRU achieved high and balanced performance (AUC-ROC 0.98), confirming the importance of modeling sequential dependencies in transaction flows. Both methods maintained good trade-offs between Precision and Recall, avoiding the under-detection problem seen in Random Forest. Hybrid models advanced performance further: GCN–CNN slightly outperformed CNN in ranking anomalous transactions (AUC-ROC = 0.9794), while GCN–GRU achieved the best overall results across all metrics (Accuracy = 0.9470, Recall = 0.9470, AUC-ROC = 0.9807). This shows that combining structural context with temporal modeling enables detection of subtle anomalies that evolve dynamically in graph-based transaction networks. Overall, the findings highlight three in-sights: (i) ensemble methods are limited by low Recall, (ii) structural-only models underperform when temporal dynamics are ignored, and (iii) hybrid graph–sequence architectures, particularly GCN–GRU, offer the most robust and balanced solution for anomaly detection in cryptocurrency transactions.

**Table 3.** Performance comparison

| Model | Accuracy | Precision | Recall | F1-score | AUC-ROC |
|---|---|---|---|---|---|
| RandomForest | 0.9343 | 0.7999 | 0.6872 | 0.7393 | 0.9607 |
| GCN | 0.7735 | 0.7996 | 0.7735 | 0.7844 | 0.6743 |
| CNN | 0.9382 | 0.9409 | 0.9382 | 0.9393 | 0.9773 |
| GCN–CNN | 0.9360 | 0.9416 | 0.9360 | 0.9380 | 0.9794 |
| GRU | 0.9360 | 0.9404 | 0.9360 | 0.9377 | 0.9786 |
| **GCN–GRU** | **0.9470** | **0.9478** | **0.9470** | **0.9474** | **0.9807** |

## 5 Discussion

### 5.1 Key Findings & Contributions

This study demonstrates that integrating structural (graph-based) and temporal (sequence-based) learning significantly improves anomaly detection in blockchain transaction networks. The proposed hybrid GCN-GRU architecture, building on the GCN formulation by Kipf Welling (2017) [13] and the GRU design by Cho et al. (2014) [14] effectively captures:

1. Topological dependencies - anomalous patterns embedded in transaction network structures.
2. Sequential dynamics - suspicious behaviors evolving over time.

Experimental results indicate that the hybrid model outperforms unimodal deep learning models (GCN, CNN, GRU) and ensemble methods in accuracy, recall, F1-score, and AUC-ROC, particularly in cases where temporal irregularities align with unusual structural patterns. Our contributions are threefold:

1. conducting a comprehensive comparative analysis against diverse baselines.
2. validating the approach on real-world Wasabi Wallet mixing transaction data, achieving robust performance despite class imbalance.
3. framing anomaly detection in cryptocurrency transaction networks as a binary classification task, one of the first attempts in this domain.

### 5.2 Limitations & Future Works

Despite its strong performance, the proposed method faces several challenges: computational cost in handling large-scale blockchain graphs with fine temporal granularity, potential difficulty in detecting fraud patterns unseen during training, and limited interpretability of model decisions. Future research will focus on

1. enhancing adaptability to novel anomalies through few-shot learning and domain adaptation.
2. integrating explainable AI (XAI) techniques to provide clearer insights for fraud investigators.

## 6 Conclusion

This study investigated the effectiveness of a hybrid deep learning approach that integrates both structural and temporal information for anomaly detection in cryptocurrency transactions. While traditional ensemble methods can improve performance by combining multiple weak classifiers, they generally fail to explicitly capture network structures or sequential dependencies. To address this limitation, we proposed a model that leverages both graph-based relational features and time-series behavioral patterns. Experimental results demonstrated that the hybrid model consistently outperformed baseline models, including traditional machine learning and unimodal deep learning architectures, in detecting illicit transactions. The findings highlight the importance of jointly modeling structural dependencies and temporal dynamics for improving anomaly detection performance in complex transactional networks. The contributions of this work include:

1. providing empirical evidence on the benefits of combining graph and sequence modeling for fraud detection.
2. demonstrating the practical applicability of hybrid models in tracking money flows, transaction networks, and anomalous transaction patterns.
3. offering a reproducible framework for future research. Future work will focus on extending the proposed architecture to other domains with similar network and temporal characteristics, as well as exploring scalability and real-time inference capabilities.

## Acknowledgments

This work was supported by the National Research Foundation of Korea (NRF) grant funded by the Korean government under the project "Socio-Technological Solutions for Bridging the AI Divide: A Blockchain and Federated Learning-Based AI Training Data Platform" (NRF-2024S1A5C3A02043653).